\pdfoutput=1

\documentclass[11pt]{article}

\usepackage[]{EMNLP2023}

\usepackage{times}
\usepackage{latexsym}

\usepackage[T1]{fontenc}

\usepackage[utf8]{inputenc}

\usepackage{microtype}

\usepackage{inconsolata}
\usepackage{graphicx} 
\usepackage{multirow}
\usepackage{booktabs}
\usepackage{amsmath}
\usepackage{arydshln}
\usepackage{colortbl}
\usepackage{hhline}
\usepackage{amssymb}
\usepackage{pifont}
\usepackage{hyperref}
\newcommand{\xmark}{\ding{55}}%

\newcommand{\ourdataset}{\textsc{heap}}
\newcommand{\grayc}[1]{\textcolor{gray}{#1}}
\newcommand{\grayline}{\arrayrulecolor{gray}\hline\arrayrulecolor{black}}

\title{Automatic Evaluation of Generative Models with Instruction Tuning}

\author{Shuhaib Mehri$^{1}$ and Vered Shwartz$^{1,2}$ \\
$^1$ University of British Columbia\\
$^2$ Vector Institute for AI\\
{\tt shuhaibm@student.ubc.ca, vshwartz@cs.ubc.ca }\\}

\begin{document}
\maketitle

\begin{abstract}
Automatic evaluation of natural language generation has long been an elusive goal in NLP. A recent paradigm fine-tunes pre-trained language models to emulate human judgements for a  particular task and evaluation criterion. Inspired by the generalization ability of instruction-tuned models, we propose a learned metric based on instruction tuning. To test our approach, we collected \ourdataset{}, a dataset of human judgements across various NLG tasks and evaluation criteria. Our findings demonstrate that instruction tuning language models on \ourdataset{} yields good performance on many evaluation tasks, though some criteria are less trivial to learn than others. Further, jointly training on multiple tasks can yield additional performance improvements, which can be beneficial for future tasks with little to no human annotated data.
\end{abstract}

\section{Introduction}
\label{sec:intro}

Natural language generation (NLG) has made significant leaps forward in recent years thanks to large language models \cite[LLMs;][]{NEURIPS2020_1457c0d6,open2023introducing}. Yet, to date, there is no standard evaluation protocol for NLG systems. Human evaluation provides the most accurate assessment, but its costly and time-consuming nature makes it less practical for large-scale evaluations, and it's rarely conducted as part of the system development cycle. For this reason, automatic evaluation metrics have been widely adopted. The majority of automatic metrics compare the system outputs against a set of reference texts, measuring either lexical overlap \cite[e.g.,][]{papineni-etal-2002-bleu,lin-2004-rouge} or semantic similarity \cite[e.g.,][]{zhang2019bertscore}. 

Reference-based metrics suffer from many drawbacks. First, system outputs that are different from the references are scored low, even if they are correct. Second, multiple studies have noted poor correlation with human judgements \cite{novikova-etal-2017-need, dhingra-etal-2019-handling,chen-etal-2019-evaluating,kryscinski-etal-2019-neural}. Third, methods that were designed with one task in mind, such as BLEU \cite{papineni-etal-2002-bleu} for machine translation and ROUGE for summarization \cite{lin-2004-rouge}, don't necessarily transfer well to other tasks  \cite{liu-etal-2016-evaluate,nema-khapra-2018-towards}. Finally, by producing a single score based on similarity to the references, some important but more nuanced dimensions might be missed, such as faithfulness, answerability, and more. 

\begin{figure}[t]
\centering
\includegraphics[width=.49\textwidth]{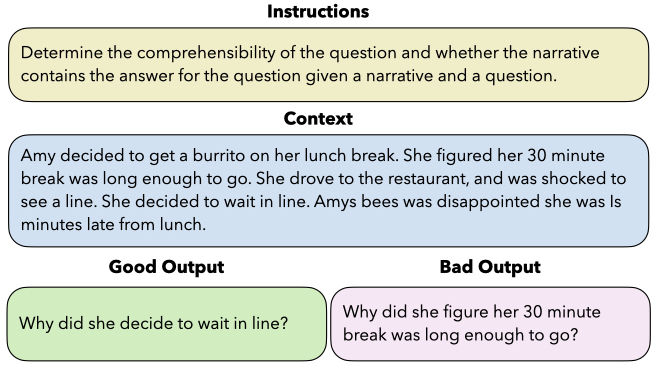}
\caption{Example from \ourdataset{}, originally taken the TellMeWhy dataset \cite{lal-etal-2021-tellmewhy}, here focusing on the question answerability (QA) criteria.}
\label{fig:example}
\end{figure}

A recent alternative approach is learned metrics. Such metrics leverage a pre-trained language model and fine-tune it to emulate human judgements \cite[e.g.;][]{sellam-etal-2020-bleurt,zhao-etal-2020-designing}. Learned metrics are typically tailored to specific tasks (e.g., machine translation) and criteria (e.g., similarity to the references), and they can be reference-based or reference-less. 

In this work, we propose to train reference-less learned metrics using instruction tuning. Instruction tuning involves presenting the model with natural language instructions in addition to the task inputs. Including the instructions as part of the input enables models to generalize better, perform well in zero-shot and few-shot settings \cite{wei2021finetuned,gupta-etal-2022-instructdial}, and better align with human values \cite{peng2023instruction}. 


To train our metric, we collected the \textbf{H}uman \textbf{E}valuations of \textbf{A}nswer \textbf{P}airs dataset (\ourdataset{}). \ourdataset{} was composed from the human evaluation results for 8 datasets, along 22 diverse evaluation criteria, such as comprehensibility, appropriateness, grammaticality, and informativeness, as detailed in Table~\ref{tab:tasks}.\footnote{The annotations were generously shared with us by the dataset creators.}  We converted all data points to a uniform comparative format, consisting of the task instructions, and two context-generation pairs, such that one generation (good in Figure~\ref{fig:example}) was ranked better than the other (bad). 

We used \ourdataset{} with instruction tuning in single-task, multi-task, and cross-task setups. We find that most criteria are learnable, though more nuanced or complex ones (e.g., answer validity) are more difficult to learn than others (e.g., grammaticality). We also show that fine-tuning on the task is essential, and that multi-tasking can help with the more difficult tasks. Finally, the cross-task setup is less successful, but can be improved by training only on a subset of similar tasks to the target task.

We hope that our findings will guide future research on automatic evaluation for NLG systems.\footnote{Code and data available at \href{https://github.com/Shuhaibm/heap}{https://github.com/Shuhaibm/heap}} 


\section{Related Work}
\label{sec:related_work}
\begin{table*}[t]
\centering
\small
\setlength{\tabcolsep}{4pt}
\begin{tabular}{lrp{8cm}}
\toprule
\textbf{Task} & \multicolumn{1}{l}{\textbf{\#Examples}} & \textbf{Dataset} \\
\midrule
Advice Helpfulness [AH] & 1,200 & TuringAdvice: Advice Generation \cite{zellers-etal-2021-turingadvice} \\
\midrule
Answer Grammaticality [AG]  & 598 & \multirow{3}{8cm}{TellMeWhy: Answering Why-Questions \cite{lal-etal-2021-tellmewhy}} \\ 
Answer Validity [AV] & 598 &  \\ 
Question Answerability [QA] & 1,917 & 
\\
\midrule
HellaSwag [HS] & 149,841 & 
HellaSwag: Commonsense NLI \cite{zellers-etal-2019-hellaswag} \\
\midrule
Commonsense Reasoning [CR] & 1,079 & CommonGen: Commonsense Reasoning \cite{lin-etal-2020-commongen} \\
\midrule
Best Counter Narrative [BCN] & 1,000 & 
\multirow{5}{8cm}{Counter Narratives Against Hate Speech \cite{tekiroglu-etal-2022-using}} \\ 
Choose-or-not [CCN] & 884 & \\ 
Grammaticality [CNG] & 863 & \\ 
Specificity [CNSp] & 1,139 &  \\ 
Suitability [CNSu] & 1,471 &  
\\
\midrule 
Counter Narrative Informativeness [CNI] & 783 & 
\multirow{4}{8cm}{CHASM: Countering Online Hate Speech and Microaggressions \cite{ashida-komachi-2022-towards}} \\ 
Counter Narrative Offensiveness [CNO] & 685 &  \\ 
Counter Narrative Stance [CNSt] & 724 &  \\ Hate Speech Offensiveness [HSO] & 29,970 & 
\\
\midrule
Story Rewriting Counterfactual [SRC] & 4,400 &
\multirow{5}{8cm}{TimeTravel: Counterfactual Story Rewriting \cite{qin-etal-2019-counterfactual}} \\ 
Story Rewriting Ending [SRE] & 4,400 &  \\ 
Story Rewriting Plot [SRPl] & 4,400 &  \\ 
Story Rewriting Premise [SRPr] & 4,400 &  \\ Story Rewriting Second [SRS] & 4,400 & 
\\
\midrule
Attenuator Effectiveness [DIA] & 7,176 &
\multirow{2}{8cm}{Defeasible Inference \cite{rudinger-etal-2020-thinking}} \\ 
Intensifier Effectiveness [DII] & 7,176 
\\
\bottomrule
\end{tabular}
\caption{Human evaluation criteria (referred to as ``tasks'' in this paper) included in \ourdataset{}.}
\label{tab:tasks}
\end{table*}

\paragraph{Automatic Evaluation of Generative Tasks.} Numerous automatic methods exist for evaluating generative models. The majority of  metrics involve assessing the similarity between a generated output and a reference text. Commonly used metrics include BLEU \cite{papineni-etal-2002-bleu}, ROUGE \cite{lin-2004-rouge}, and METEOR \cite{banerjee-lavie-2005-meteor}, which focus on measuring lexical overlap between generated outputs and a reference. More recent methods, such as BERTScore \cite{zhang2019bertscore}, go beyond lexical overlap by embedding both the generations and the references into a shared space and computing cosine similarity between the embeddings. All these metrics operate at the surface level, predominantly focusing on lexical similarity.

Some metrics have been proposed which are trained to emulate human judgements. BLEURT \cite{sellam-etal-2020-bleurt} is a BERT-based metric which is first trained to estimate the scores from existing automatic metrics for a large number of synthetic sentence pairs, and then trained to emulate human judgements for a machine translation task. Similarly,  \newcite{zhao-etal-2020-designing} proposed a RoBERTa-based metric for dialogue evaluation which is first trained on a large number of sentence pairs with a next sentence prediction objective, and then trained on a small number of human annotations for the task. Learned metrics are tailored to specific tasks. They can take on different forms: reference-based, where the metric is trained to compare the system's output to a reference text, like in BLUERT; reference-less, where the metric scores the output along some criterion without the use of references \cite{sinha-etal-2020-learning}; or a combination of both, as seen in \newcite{ghazarian-etal-2019-better}'s work. In this work, we propose a reference-less learned metric and investigate the transferability between different tasks and criteria.

\paragraph{Instruction Tuning.} Instruction tuning is a fine-tuning technique that involves training a model on a variety of tasks, leveraging natural language instructions to guide the model towards producing the correct answers. Recent studies have showcased the effectiveness of this technique in improving LLMs' ability to generalize in a zero-shot and few-shot setting \cite{chen-etal-2022-meta, wei2021finetuned, peng2023instruction}. Most pertinent to our work, \newcite{gupta-etal-2022-instructdial} applied instruction tuning to 48 dialogue-related tasks, including dialogue evaluation. They showed that their instruction-tuned dialogue evaluation metric achieves improved correlation with human judgements, even in a cross-task setup when training on other dialogue tasks. In this work, we use instruction-tuning to train automatic evaluation metrics for a diverse set of tasks and criteria. The use of instructions allows for more transferability between different tasks and criteria, and could be beneficial when data for a particular task is sparse. 

\section{Dataset}
\label{sec:data}
We introduce the \textbf{H}uman \textbf{E}valuations of \textbf{A}nswer \textbf{P}airs (\ourdataset{}) dataset. \ourdataset{} is designed to train and evaluate automatic methods for the evaluation of generative tasks. It is derived from existing human evaluations that were performed on 8 generative tasks detailed in Table~\ref{tab:tasks}. We obtained the data from public releases as well as by reaching out to the authors of the respective papers. Appendix~\ref{appendix:instructions} provides examples for each task and criteria along with the instructions we used for them.

The original human evaluations for some of the datasets included comparative evaluation (i.e., which of the answers is better along some criterion), while others included absolute scores of an answer's quality. We decided to go with the comparative setup based on the findings of \newcite{askell2021general} and \newcite{bai2022training} who demonstrated that a ranked preference model, which is a model trained to assign a higher score to the `better' sample in a given pair, outperforms other training objectives like imitation learning and binary discrimination. To that end, we converted absolute scores and comparison between multiple answers into pairwise comparisons. 

The dataset contains 229,104 instances. The instances from each task are randomly split into 80\% train, 10\% validation, and 10\% test sets and combined. Each data point in \ourdataset{} consists of two generated outputs, \texttt{good\_sample} and \texttt{bad\_sample}, where each sample has its own context \texttt{C}. Each data point belongs to a ``task'', which is a combination of the original dataset (e.g. advice generation) and evaluation criterion (e.g. advice helpfulness). An example data point can be seen in Figure~\ref{fig:example}.

\section{Method}
\label{sec:method}
We propose to fine-tune pre-trained language models to predict a scalar score for text outputs along various criteria. We train the models using natural language instructions (Sec~\ref{sec:instruction_tuning}) and investigate the extent that this setup allows for out-of-domain generalization for new tasks (Sec~\ref{sec:eval_setups}). 

\subsection{Instruction Tuning}
\label{sec:instruction_tuning}

Instruction tuning refers to a setup in which natural language instructions are prepended to the input (Figure~\ref{fig:example}). By incorporating instructions in a model's training, it learns how to arrive at the expected output for a given task \cite{mishra-etal-2022-cross}. 

To find the optimal instructions for each task, we manually wrote a diverse set of instructions and chose the instruction that yielded the best performance on the task's validation set. Details about the instructions used can be found in Appendix~\ref{appendix:instructions}.


We used BART-base \cite{lewis-etal-2020-bart}, a pre-trained encoder-decoder model with 140M parameters, for all our experiments. We fine-tuned BART to predict a score for each answer. Specifically, the input for each example is in the following format: \texttt{<instructions> <context> <answer>}. We embed the input using BART and feed the last hidden state into a linear layer to predict a scalar score $r$, where a higher score denotes a more favourable input. Following prior work \cite{NIPS2017_d5e2c0ad,askell2021general}, we maximize the difference between the scores of the good and bad outputs with the following loss function: $\mathcal{L} = \log(1 + \exp(r_{\mathrm{bad}} - r_{\mathrm{good}}))$.


\subsection{Evaluation Setups}
\label{sec:eval_setups}

We train and evaluate the models in the following setups: 

\paragraph{Single-Task.} In this setup, for a target task $t$, we train the model on the training set composed of only $t$'s instances ($D_{\operatorname{train}}^t$) and test it on the test set composed of only $t$'s instances ($D_{\operatorname{test}}^t$). 

\paragraph{Multi-Task.} In this setup, we investigate whether the different tasks can mutually benefit each other. We train a single model on the entire \ourdataset{} training set ($D_{\operatorname{train}}$) and test it on the test set composed of only $t$'s instances ($D_{\operatorname{test}}^t$). 

\paragraph{Cross-Task.} In this setup, we investigate our instruction-tuned models' zero-shot generalization abilities, by evaluating them on unseen tasks. For a target task $t$, we train a model on the \ourdataset{} training set excluding $t$'s instances ($D_{\operatorname{train}}^{/t}$), and test it on the test set composed of only $t$'s instances ($D_{\operatorname{test}}^t$). We hypothesized that the model would be able to generalize to a new task by learning to follow instructions. 

\paragraph{Cross-Cluster.} In this setup, we repeat the cross-task setup, but train the model on a subset of \ourdataset{}. We refer to each such subset as a ``cluster''. Each cluster consists of handpicked tasks based on certain similarities. For a target task $t$ that belongs to cluster $C$, we train a model on the cluster's training set excluding $t$'s instances ($C_{\operatorname{train}}^{/t}$), and test it on the test set composed of only $t$'s instances ($C_{\operatorname{test}}^t$). We hypothesize that being more selective with tasks will further improve a model's ability to generalize to a new task.


\section{Experimental Setup}
\label{sec:exp_setup}
\begin{table*}
\centering
\small
\setlength{\tabcolsep}{5pt}
\begin{tabular}{l|rrrr|rrrr}
\toprule
\textbf{Task} & \multicolumn{4}{c}{\textbf{Accuracy}} &  \multicolumn{4}{c}{\textbf{Spearman Rank-order Correlation}} \\
\midrule
& \multicolumn{1}{c}{\textbf{Base}} & \multicolumn{1}{c}{\textbf{Single-task}} & \multicolumn{1}{c}{\textbf{Multi-task}} & \multicolumn{1}{c}{\textbf{Cross-task}} & \multicolumn{1}{c}{\textbf{Base}} & \multicolumn{1}{c}{\textbf{Single-task}} & \multicolumn{1}{c}{\textbf{Multi-task}} & \multicolumn{1}{c}{\textbf{Cross-task}} \\
\midrule
AH & 52.55$\pm$8.02  & \textbf{67.33$\pm$1.60}  & \underline{66.94$\pm$1.27}  & 55.84$\pm$7.22  & \multicolumn{1}{c}{-} & \multicolumn{1}{c}{-} & \multicolumn{1}{c}{-} & \multicolumn{1}{c}{-} \\

AG   & 58.97$\pm$14.66  & \underline{69.47$\pm$5.46}  & \textbf{78.07$\pm$6.62}  & 58.12$\pm$3.91 & 0.181 & \textbf{0.379} & \underline{0.231} & 0.146 \\

AV    & 53.9$\pm$2.66  & 44.35$\pm$10.03  & \textbf{65.22$\pm$5.75}  & \underline{58.87$\pm$10.92}  & 0.063 & -0.026
& \textbf{0.293} & \underline{0.093} \\

QA  & 46.35$\pm$14.11  & \underline{58.96$\pm$14.58}  & \textbf{75.17$\pm$2.10}  & 42.19$\pm$3.76  & \multicolumn{1}{c}{-} & \multicolumn{1}{c}{-} & \multicolumn{1}{c}{-} & \multicolumn{1}{c}{-} \\

HS     & 49.76$\pm$0.62  & \textbf{67.91$\pm$0.42}  & \underline{65.62$\pm$0.45}  & 51.19$\pm$0.10  & \multicolumn{1}{c}{-} & \multicolumn{1}{c}{-} & \multicolumn{1}{c}{-} & \multicolumn{1}{c}{-} \\

CR   & 49.07$\pm$5.29  & \underline{77.96$\pm$4.37}  & \textbf{79.32$\pm$1.07}  & 54.32$\pm$3.86 & \multicolumn{1}{c}{-} & \multicolumn{1}{c}{-} & \multicolumn{1}{c}{-} & \multicolumn{1}{c}{-} \\

BCN  & 47.33$\pm$2.05  & 50.60$\pm$3.29  & \textbf{79.32$\pm$1.73}  & \underline{63.00$\pm$2.00} & \multicolumn{1}{c}{-} & \multicolumn{1}{c}{-} & \multicolumn{1}{c}{-} & \multicolumn{1}{c}{-} \\

CCN   & 44.19$\pm$4.62  & 64.54$\pm$12.48  & \textbf{68.18$\pm$1.14}  & \underline{68.16$\pm$1.72} & \multicolumn{1}{c}{-} & \multicolumn{1}{c}{-} & \multicolumn{1}{c}{-} & \multicolumn{1}{c}{-} \\

CNG  & 56.32$\pm$2.48  & \underline{77.67$\pm$2.24}  & \textbf{82.17$\pm$1.78}  & 62.84$\pm$2.89 & 0.089 & \underline{0.472} & \textbf{0.538} & 0.193 \\

CNSp  & 51.17$\pm$3.94  & \underline{54.39$\pm$4.85}  & \textbf{64.03$\pm$3.16}  & 48.54$\pm$3.95  & 0.086 & \underline{0.211} & \textbf{0.278} & 0.090 \\

CNSu  & 43.24$\pm$5.06  & 58.64$\pm$6.23  & \textbf{68.03$\pm$2.45}  & \underline{64.64$\pm$1.03} & -0.152 & \underline{0.143} & \textbf{0.413} & 0.081 \\

CNI  & 41.77$\pm$5.76  & \textbf{83.59$\pm$2.11}  & \underline{76.92$\pm$3.39}  & 70.88$\pm$4.39 & -0.081 & \textbf{0.574} & \underline{0.472} & -0.014 \\

CNO  & 55.07$\pm$9.47  & \underline{67.35$\pm$6.19}  & \textbf{69.12$\pm$1.47}  & 28.50$\pm$3.64  & 0.181 & \underline{0.440}
& \textbf{0.580} & -0.125 \\

CNSt  & 47.04$\pm$3.92  & \textbf{76.39$\pm$2.78}  & \underline{71.30$\pm$5.61}  & 51.60$\pm$4.18  & -0.087 & \underline{0.436} & \textbf{0.461} & 0.129 \\

HSO  & 46.15$\pm$7.72  & \textbf{68.81$\pm$2.94}  & \underline{66.43$\pm$2.53}  & 49.55$\pm$0.76  & -0.170 & \textbf{0.425} & \underline{0.399} & 0.086 \\

SRC  & 44.81$\pm$3.37  & \underline{51.80$\pm$2.74}  & \textbf{57.38$\pm$1.64}  & 48.09$\pm$6.21  & \multicolumn{1}{c}{-} & \multicolumn{1}{c}{-} & \multicolumn{1}{c}{-} & \multicolumn{1}{c}{-} \\

SRE  & 37.5$\pm$6.36  & 49.64$\pm$9.81  & \textbf{58.93$\pm$5.36}  & \underline{56.55$\pm$5.16} & \multicolumn{1}{c}{-} & \multicolumn{1}{c}{-} & \multicolumn{1}{c}{-} & \multicolumn{1}{c}{-} \\

SRPl  & 45.89$\pm$20.37  & \underline{70.72$\pm$1.21}  & \textbf{75.36$\pm$6.31}  & 50.24$\pm$6.03 & \multicolumn{1}{c}{-} & \multicolumn{1}{c}{-} & \multicolumn{1}{c}{-} & \multicolumn{1}{c}{-} \\

SRPr  & 40.67$\pm$5.73  & \underline{49.60$\pm$7.40}  & \textbf{56.00$\pm$3.46}  & 44.67$\pm$3.06 & \multicolumn{1}{c}{-} & \multicolumn{1}{c}{-} & \multicolumn{1}{c}{-} & \multicolumn{1}{c}{-} \\

SRS  & 52.22$\pm$2.83  & \underline{55.33$\pm$7.40}  & \textbf{61.67$\pm$3.34}  & 55.00$\pm$5.00 & \multicolumn{1}{c}{-} & \multicolumn{1}{c}{-} & \multicolumn{1}{c}{-} & \multicolumn{1}{c}{-} \\

DIA  & 48.24$\pm$2.61  & \underline{65.43$\pm$6.25}  & \textbf{69.68$\pm$1.90}  & 62.03$\pm$2.29 & -0.009 & \underline{0.302} & \textbf{0.321} & 0.268 \\

DII  & 49.93$\pm$2.43  & \underline{60.82$\pm$3.65}  & 59.26$\pm$0.71  & \textbf{61.84$\pm$5.09} & 0.060 & \textbf{0.252} & \underline{0.215} & 0.080 \\
\bottomrule
\end{tabular}
\caption{Accuracy on the test set, and Spearman rank-order correlation with human judgements, for each task in each of the setups detailed in Sec~\ref{sec:eval_setups}. Accuracy is reported as the average of 5 runs with different random seeds. Correlation is reported for datasets that have ranked data. Bold indicates best performance and underline indicates second-best. \textbf{Takeaways}: (i) fine-tuning is essential; (ii) training on additional tasks is beneficial for most target tasks; (iii) success in the cross-task setup varies a lot.}
\label{tab:results}
\end{table*}

\paragraph{Baselines.} Other than the single-task, multi-task, cross-task, and cross-cluster setups described in Sec~\ref{sec:eval_setups}, we also included the \textbf{base} setup, in which we used BART off-the-shelf without fine-tuning it.

\paragraph{Hyper-parameter Tuning.} We performed hyper-parameter tuning on the validation set to select values for the following: learning rate ($2e-5$, $2e-4$, $3e-4$), gradient accumulation ($4$, $8$, $16$, $32$, $64$, $128$), number of epochs ($1-20$), truncation, instructions, and labelling elements of the input. The selected values are available in Appendix~\ref{appendix:hyperparams}.

\paragraph{Evaluation Metrics.} We evaluated the performance of our models using two metrics. The first metric is accuracy on the respective test set. That is, we obtained scores $r_{good}$ and $r_{bad}$ for the respective answers, and counted the percent of instances for which $r_{good}$ was greater than $r_{bad}$. The second metric is Spearman rank-order correlation between the scores outputted by the model and the original human evaluation scores. This metric shows the extent to which the model's preferences align with human preferences.

\section{Results}
\label{sec:results}
Table~\ref{tab:results} presents the main results. We observe the following. 

\paragraph{Fine-tuning is essential.} The base model has an average accuracy of 50.58\% across tasks, which is akin to a random baseline. The single task setup substantially improves upon the base model with an average of 63.24\%. 

\paragraph{Most criteria are learnable,} as evident by the 12.66\% difference in accuracy between the base and the single-task models. However, for a few tasks, even the best performance remains relatively low: SRC, SRE, SRPr, and SRS. These tasks all come from the TimeTravel dataset of counterfactual story rewriting \cite{qin-etal-2019-counterfactual} and they are inherently difficult, as they require comparing two almost identical stories along various dimensions.


\paragraph{Multi-tasking is beneficial.} On average, the multi-task setup achieves 68.82\% accuracy, 5.58\% higher than the single-task setup. This indicates that there is transfer learning among the different tasks. Perhaps trivially, multi-tasking is especially beneficial when the single-task accuracy is low. The performance of tasks such as CNSt, CNG, CR, and CNI that already achieve good performance in the single-task setup either decreases or increases very slightly. Conversely, multi-tasking is the most beneficial for tasks that achieve low single-task performance, such as AV and BCN. 

\paragraph{Success in the cross-task setup varies.} The cross-task setup performs substantially worse than the multi-task setup (54.85\% compared to 68.82\% on average), which is expected since the target task training data is excluded. Compared to the single-task setup, the cross-task setup is beneficial for CCN, CNSu, SRE, BCN, and AV, but even in those cases, it is less beneficial than the multi-task setup. For CNO, the cross-setup performed substantially worse even than the baseline, but we couldn't find a reasonable explanation for this behavior.


\paragraph{The number of per-task examples is not the most important factor.} Notably, the number of examples available for each task had very weak correlations with the single-task performance (Pearson $\rho= 0.15$), the gain from multi-tasking ($\rho= -0.25$), and the gain from the cross-task setup ($\rho= 0.16$). We conclude that among the important indicators for good performance are the ease of the task, i.e., ``easy'' tasks such as verifying grammaticality can already achieve good performance without training on additional tasks. 


\definecolor{maroon}{cmyk}{0,0.87,0.68,0.32}

\begin{table}[t]
\centering
\small
\setlength{\tabcolsep}{4pt}
\begin{tabular}{lrrr}
\toprule
\textbf{Task} & \multicolumn{1}{c}{\textbf{Single-Task}} & \multicolumn{1}{c}{\textbf{Cross All}} & \multicolumn{1}{c}{\textbf{Cross Cluster}} \\
\midrule
\rowcolor{maroon!30}
\multicolumn{4}{l}{\textbf{Cluster 1: Require understanding the context C}} \\
\midrule
AH & \grayc{67.33$\pm$1.60} & 55.84$\pm$7.22 & \textbf{56.39$\pm$2.08} \\
AV & \grayc{44.35$\pm$10.03} & \textbf{58.87$\pm$10.92} & 53.9$\pm$1.00 \\
QA & \grayc{58.96$\pm$14.58} & 42.19$\pm$3.76  & \textbf{52.78$\pm$5.79} \\
HS & \grayc{67.91$\pm$0.42} & \textbf{51.19$\pm$0.10} & 44.46$\pm$0.44 \\
CR & \grayc{77.96$\pm$4.37} & \textbf{54.32$\pm$3.86} & 48.77$\pm$1.90 \\
BCN & \grayc{50.60$\pm$3.29} & \textbf{63.00$\pm$2.00} & 58.00$\pm$2.16 \\
SRC & \grayc{51.80$\pm$2.74} & \textbf{48.09$\pm$6.21} & 41.53$\pm$4.70\\
DIA & \grayc{65.43$\pm$6.25} & \textbf{62.03$\pm$2.29} & 43.86$\pm$2.57 \\
DII & \grayc{60.82$\pm$3.65} & \textbf{61.84$\pm$5.09} & 50.61$\pm$4.45 \\
\grayline
\textbf{Average} & - & 55.26 & 50.03 \\
\midrule
\rowcolor{maroon!30} \multicolumn{4}{l}{\textbf{Cluster 2: Don't require understanding the context C}} \\
\midrule
AG & \grayc{69.47$\pm$5.46} & 58.12$\pm$3.91 & \textbf{65.81$\pm$8.46} \\
CNG & \grayc{77.67$\pm$2.24} & \textbf{62.84$\pm$2.89} & 56.32$\pm$1.63 \\
CNSp & \grayc{54.39$\pm$4.85} & 48.54$\pm$3.95 & \textbf{51.17$\pm$1.49} \\
CNSu & \grayc{58.64$\pm$6.23} & \textbf{64.64$\pm$1.03} & 54.96$\pm$3.04 \\
CNI & \grayc{83.59$\pm$2.11} & \textbf{70.88$\pm$4.39} & 54.85$\pm$5.31 \\
CNO & \grayc{67.35$\pm$6.19} & 28.50$\pm$3.64 & \textbf{43.96$\pm$3.62} \\
CNSt & \grayc{76.39$\pm$2.78} & 51.60$\pm$4.18 & \textbf{65.75$\pm$3.87} \\
HSO & \grayc{68.81$\pm$2.94} & 49.55$\pm$0.76 & \textbf{58.06$\pm$3.48} \\
SRPl & \grayc{70.72$\pm$1.21} & \textbf{50.24$\pm$6.03} & 47.83$\pm$1.18 \\
SRPr & \grayc{49.60$\pm$7.40} & 44.67$\pm$3.06 & \textbf{58.00$\pm$5.89} \\
\grayline
\textbf{Average} & - & 52.96 & 55.67 \\
\midrule
\rowcolor{maroon!30} \multicolumn{4}{l}{\textbf{Cluster 3: Hate speech related tasks}} \\
\midrule
BCN & \grayc{50.60$\pm$3.29} & 63.00$\pm$2.00 & \textbf{69.00$\pm$4.58} \\
CCN & \grayc{64.54$\pm$12.48} & \textbf{68.16$\pm$1.72} & 64.05$\pm$4.05 \\
CNG & \grayc{77.67$\pm$2.24} & 62.84$\pm$2.89 & \textbf{63.22$\pm$4.6} \\ 
CNSp & \grayc{54.39$\pm$4.85} & 48.54$\pm$3.95 & \textbf{59.36$\pm$0.51} \\
CNSu & \grayc{58.64$\pm$6.23} & \textbf{64.64$\pm$1.03} & \textbf{64.64$\pm$2.17} \\
CNI & \grayc{83.59$\pm$2.11} & \textbf{70.88$\pm$4.39} & 57.00$\pm$2.47 \\
CNO & \grayc{67.35$\pm$6.19} & 28.50$\pm$3.64 & \textbf{45.38$\pm$4.68}\\
CNSt & \grayc{76.39$\pm$2.78} & 51.60$\pm$4.18 & \textbf{67.58$\pm$2.85} \\
HSO & \grayc{68.81$\pm$2.94} & 49.55$\pm$0.76 & \textbf{52.50$\pm$1.48} \\
\grayline
\textbf{Average} & - & 56.41 & 60.30 \\
\bottomrule
\end{tabular}
\caption{Per-task accuracy when the model is trained on all other tasks in the cross-task setup (\textbf{Cross All}) vs. all other tasks in the same cluster (\textbf{Cross Cluster}).}
\label{tab:cluster}
\end{table}

\paragraph{Choosing the right tasks for transfer matters.} Results for the cross-cluster setup is presented in Table~\ref{tab:cluster}. The unsurprising finding is that one can benefit from training on a cluster that consists of similar tasks. For example, the first cluster consists of tasks that require deep semantic understanding of the context C. The tasks in this cluster are diverse, ranging from advice helpfulness through general commonsense reasoning to defeasible and counterfactual reasoning. As a result, the average accuracy for the tasks in this cluster drops from 55.26\% to 50.03\%.

Conversely, when the clusters involve more closely-related tasks, it is beneficial to limit the training to the cluster tasks. For example, tasks that require more superficial  understanding of the context C or none at all, involve evaluating the grammaticality, specificity, suitability, informativeness, offensiveness, and stance of the generated answers. Those tasks are related enough to increase the average accuracy from 52.96\% to 55.67\%. When further focusing on tasks coming from similar datasets, such as tasks pertaining to hate speech detection, the performance improvement is more substantial (56.41\% to 60.3\%). A similar trend holds when focusing on different criteria from the same dataset, e.g. from 48.66\% to 53.06\% on answering why-questions, 60.56\% to 61.94\% on defeasible inference, and 50.91\% to 53.22 on counterfactual story rewriting.



\section{Conclusion}
\label{sec:conclusion}
We proposed to use instruction tuning to learn automatic evaluation metrics. To test the effectiveness of this approach, we introduced \ourdataset{}, a collection of human judgements along diverse dimensions for various generative tasks. Our experiments confirm the importance of fine-tuning for developing metrics that align with human judgements. Further, we showed the advantage of fine-tuning on multiple tasks, and that a cross-task (zero-shot) setup yields positive results when trained on selected tasks. Collectively, our experiments reveal the value of instruction tuning in the domain of automatic evaluation of generative tasks. We hope that our findings will serve as a catalyst for inspiring future research on this topic.

\section*{Limitations}
\label{sec:limitations}
\paragraph{Task Balance.} The number of examples in \ourdataset{} is imbalanced across tasks, as can be seen in Table~\ref{tab:tasks}. The number of examples range from 598 for AG and AV to 149,841 for HS. In preliminary experiments we tried to obtain a more balanced dataset by removing HS from cluster 1 (Table~\ref{tab:cluster}). This resulted in a drop of one point in average accuracy, but a significantly shorter training time. In the future, we will explore the possibility of obtaining more annotations for ``lower-resource'' tasks, applying data augmentation methods, or using more sophisticated multi-tasking techniques to overcome task imbalance.  

\paragraph{Inherent Subjectivity.} Our dataset is based on annotators' judgements of model-generated outputs along various dimensions. It's possible that some tasks involve inherent subjectivity, thus creating variance in the quality and consistency of the data for different tasks. This could further explain why our models were able to achieve better performance on more objective tasks, such as grammaticality judgement (Sec~\ref{sec:results}). 


\section*{Ethics Statement}
\label{sec:ethics}
\paragraph{Data.} The \ourdataset{} dataset is a compilation of human evaluations. We obtained them from public releases as well as by reaching out to the authors of the dataset papers. We plan to make it publicly available with the consent of the authors that contributed data. The annotations in the dataset do not include any personal information of the annotators. Details about the compensation for the annotators is available in the original papers. Finally, the contexts in \ourdataset{} come from diverse datasets (Table~\ref{tab:tasks}), some of which may include offensive, hateful, or sexual content. We did not perform quality control beyond what was performed by the original dataset creators. 

\paragraph{Models.} The \ourdataset{} dataset contains human judgements along various tasks, which may exhibit societal biases. Given that our evaluation models are trained to emulate these human judgements, it is possible that our models replicate these undesired biases.

\section*{Acknowledgements}
\label{sec:acknowledgements}
We are sincerely thankful for the authors of the datasets used in this paper for sharing the human evaluation results with us. This work was funded, in part, by the Vector Institute for AI, Canada CIFAR AI Chairs program, an NSERC discovery grant, and a research gift from AI2.

\bibliography{anthology,custom}
\bibliographystyle{acl_natbib}

\newpage
\appendix

\section{Hyper-Parameters}
\label{appendix:hyperparams}
\definecolor{green}{rgb}{0.0, 0.5, 0.0}

\begin{table}[h]
\centering
\scriptsize
\setlength{\tabcolsep}{2pt}
\begin{tabular}{llllll}
\toprule

\textbf{Task} & \textbf{Truncate} & \textbf{Label} & \textbf{Gradient} & \textbf{Learning} & \textbf{\#Epochs}\\
& \textbf{Right} & \textbf{Input} & \textbf{Accumulation} & \textbf{Rate} \\
\midrule

\rowcolor{maroon!30}
\multicolumn{6}{c}{\textbf{Single-Task Setup}} \\

AH & \textcolor{green}{$\checkmark$} & \textcolor{red}{\xmark} & 4 & 2e-5 & 17\\

AG & \textcolor{green}{$\checkmark$} & \textcolor{green}{$\checkmark$} & 8 & 2e-5 & 15\\


AV & \textcolor{green}{$\checkmark$} & \textcolor{green}{$\checkmark$} & 32 & 3e-4 & 10\\


QA & \textcolor{green}{$\checkmark$} & \textcolor{red}{\xmark} & 4 & 2e-4 & 20\\


HS & \textcolor{green}{$\checkmark$} & \textcolor{red}{\xmark} & 64 & 2e-5 & 17\\


CR & \textcolor{green}{$\checkmark$} & \textcolor{red}{\xmark} & 16 & 2e-4 & 18\\


BCN & \textcolor{green}{$\checkmark$} & \textcolor{green}{$\checkmark$} & 4 & 3e-4 & 3\\


CCN & \textcolor{green}{$\checkmark$} & \textcolor{green}{$\checkmark$} & 32 & 2e-4 & 15\\


CNG & \textcolor{green}{$\checkmark$} & \textcolor{green}{$\checkmark$} & 32 & 2e-4 & 12\\


CNSp & \textcolor{green}{$\checkmark$} & \textcolor{red}{\xmark} & 8 & 2e-4 & 12\\


CNSu & \textcolor{green}{$\checkmark$} & \textcolor{green}{$\checkmark$} & 128 & 3e-4 & 13\\


CNI & \textcolor{green}{$\checkmark$} & \textcolor{red}{\xmark} & 32 & 2e-4 & 15\\


CNO & \textcolor{green}{$\checkmark$} & \textcolor{green}{$\checkmark$} & 64 & 3e-4 & 5\\


CNSt & \textcolor{green}{$\checkmark$} & \textcolor{red}{\xmark} & 4 & 2e-5 & 13\\


HSO & \textcolor{green}{$\checkmark$} & \textcolor{green}{$\checkmark$} & 128 & 3e-4 & 2\\


SRC & \textcolor{green}{$\checkmark$} & \textcolor{green}{$\checkmark$} & 64 & 2e-4 & 7\\


SRE & \textcolor{green}{$\checkmark$} & \textcolor{green}{$\checkmark$} & 16 & 3e-4 & 2\\


SRPl & \textcolor{green}{$\checkmark$} & \textcolor{green}{$\checkmark$} & 32 & 2e-4 & 15\\


SRPr & \textcolor{green}{$\checkmark$} & \textcolor{red}{\xmark} & 8 & 3e-4 & 4\\


SRS & \textcolor{green}{$\checkmark$} & \textcolor{red}{\xmark} & 8 & 2e-5 & 18\\


DIA & \textcolor{green}{$\checkmark$} & \textcolor{red}{\xmark} & 32 & 2e-4 & 15\\


DII & \textcolor{green}{$\checkmark$} & \textcolor{red}{\xmark} & 64 & 2e-5 & 17\\

\rowcolor{maroon!30} 
\multicolumn{6}{c}{\textbf{Multi-Task Setup}} \\

Overall & - & -& 8 & 2e-5 & 16\\

\rowcolor{maroon!30} 
\multicolumn{6}{c}{\textbf{Cross-Task Setup}} \\

AH & \textcolor{green}{$\checkmark$} & \textcolor{red}{\xmark} & 4 & 2e-5 & 16\\

AG & \textcolor{green}{$\checkmark$} & \textcolor{green}{$\checkmark$} & 8 & 2e-5 & 19\\


AV & \textcolor{green}{$\checkmark$} & \textcolor{green}{$\checkmark$} & 16 & 3e-4 & 17\\


QA & \textcolor{green}{$\checkmark$} & \textcolor{red}{\xmark} & 32 & 2e-5 & 9\\


HS & \textcolor{green}{$\checkmark$} & \textcolor{red}{\xmark} & 32 & 2e-5 & 6\\


CR & \textcolor{green}{$\checkmark$} & \textcolor{red}{\xmark} & 8 & 2e-5 & 20\\


BCN & \textcolor{green}{$\checkmark$} & \textcolor{green}{$\checkmark$} & 8 & 2e-5 & 15\\


CCN & \textcolor{green}{$\checkmark$} & \textcolor{green}{$\checkmark$} & 16 & 2e-5 & 17\\


CNG & \textcolor{green}{$\checkmark$} & \textcolor{green}{$\checkmark$} & 32 & 2e-5 & 20\\


CNSp & \textcolor{green}{$\checkmark$} & \textcolor{red}{\xmark} & 16 & 2e-5 & 5\\


CNSu & \textcolor{green}{$\checkmark$} & \textcolor{green}{$\checkmark$} & 32 & 2e-5 & 15\\


CNI & \textcolor{green}{$\checkmark$} & \textcolor{red}{\xmark} & 16 & 2e-5 & 19\\


CNO & \textcolor{green}{$\checkmark$} & \textcolor{green}{$\checkmark$} & 16 & 3e-4 & 12\\


CNSt & \textcolor{green}{$\checkmark$} & \textcolor{red}{\xmark} & 32 & 2e-5 & 16\\


HSO & \textcolor{green}{$\checkmark$} & \textcolor{green}{$\checkmark$} & 32 & 3e-4 & 1\\


SRC & \textcolor{green}{$\checkmark$} & \textcolor{green}{$\checkmark$} & 4 & 2e-5 & 15\\


SRE & \textcolor{green}{$\checkmark$} & \textcolor{green}{$\checkmark$} & 32 & 2e-5 & 8\\


SRPl & \textcolor{green}{$\checkmark$} & \textcolor{green}{$\checkmark$} & 8 & 2e-5 & 1\\


SRPr & \textcolor{green}{$\checkmark$} & \textcolor{red}{\xmark} & 4 & 2e-5 & 4\\


SRS & \textcolor{green}{$\checkmark$} & \textcolor{red}{\xmark} & 32 & 2e-5 & 4\\


DIA & \textcolor{green}{$\checkmark$} & \textcolor{red}{\xmark} & 8 & 2e-5 & 14\\


DII & \textcolor{green}{$\checkmark$} & \textcolor{red}{\xmark} & 32 & 2e-5 & 18\\

\bottomrule
\end{tabular}
\caption{Hyper-paramaters used for our models.}
\label{tab:hyperparams}
\end{table}

Table~\ref{tab:hyperparams} displays the hyper-parameters used in this work. ``Label input'' refers to labeling the elements of each instance in the input, as demonstrated in Table~\ref{tab:instructions}, for example for AG.

\newpage
\section{Task Instructions}
\label{appendix:instructions}
\begin{table*}[t]
\centering
\scriptsize
\setlength{\tabcolsep}{2pt}
\begin{tabular}
{lp{0.5\linewidth}p{0.45\linewidth}}
\toprule
\textbf{Task} & \textbf{Instruction} & \textbf{Example} \\
\midrule
AH & 
Determine how helpful the advice is given a situation and advice. & 
Was summoned for Jury Duty in a state that I no longer live in and they won't approve my exclusion requests. 
[...]
I'm pretty sure I was summoned because my Driver's License is still issued (MO), but I live in VA now. 
I don't have the means to get back to MO before the jurors are decided, but I also don't want to be held in contempt of court. Anything would be helpful, thanks! \texttt{<sep>} \textcolor{blue}{Well, you are long-overdue to get your VA Driver's License (you only have 60 days), so I suggest you do so now.}\\

\midrule

AG &
Determine the grammaticality of an answer given a narrative, a question and an answer to that question. &

\multirow{6}{0.45\textwidth}{
narrative \texttt{<sep>} Tim was unemployed. He applied to many jobs. He interviewed for a car valet job. He got the job. He happily began working as a car valet. \texttt{<sep>} question \texttt{<sep>} Why did He get the job? \texttt{<sep>} answer \texttt{<sep>} \textcolor{blue}{Tim got the job because he impressed the company.}}\\

\hhline{--~}

AV & 
Determine the plausibility of the answer given a narrative, a question and an answer to that question. \\

\hhline{--~}

QA & 
Determine the comprehensibility of the question and whether the narrative contains the answer for the question given a narrative and a question.\\

\midrule

HS & 
Determine how appropriate the follow up sentence is for the context given a context and a follow up sentence. &
Then, the man writes over the snow covering the window of a car, and a woman wearing winter clothes smiles. then \texttt{<sep>} , \textcolor{blue}{the man continues removing the snow on his car.}\\

\midrule

CR & 
Given a concept set, a reference sentence, and a sentence, determine the plausibility of the sentence. &
food eat chair sit \texttt{<sep>} A man sitting on a chair eating food. \texttt{<sep>} \textcolor{blue}{a man eats food and eat chair sit in the beach.}\\

\midrule

BCN & 
Given a hate speech, the target of the hate speech, and a counter narrative in response to the hate speech, determine how good the counter narrative is. &

\multirow{12}{0.45\textwidth}{
hate speech \texttt{<sep>} blind people don't know what's going on \texttt{<sep>} target \texttt{<sep>} DISABLED \texttt{<sep>} counter narrative \texttt{<sep>} \textcolor{blue}{The problem is that blind people are not aware of what is going on, they just want to be accepted as human beings.}}\\


\hhline{--~}

CCN & 
Given a hate speech, the target of the hate speech, and a counter narrative in response to the hate speech, determine how appropriate the counter narrative is in a real case scenario.\\

\hhline{--~}

CNG & 
You are given a hate speech, the target of the hate speech, and a counter narrative in response to the hate speech. Determine the grammaticality of the counter narrative. \\

\hhline{--~}

CNSp & 
Determine how specific the arguments brought by the counter narrative are in response to the hate speech given a hate speech, the target of the hate speech, and a counter narrative in response to the hate speech.\\ 

\hhline{--~}





CNSu & 
Given a hate speech, the target of the hate speech, and a counter narrative in response to the hate speech, determine how suitable the counter narrative is to the hate speech in terms of semantic relatedness and in terms of not spreading hate. \\

\midrule

CNI & 
Determine how informative the counter narrative is given a hate speech and a counter narrative in response to the hate speech. & 
\multirow{6}{0.45\textwidth}{Some races have lower physical and cognitive abilities, the sooner we accept this, the better it will be for humanity. \texttt{<sep>} \textcolor{blue}{This is a racist statement that is not supported by science.}}\\

\hhline{--~}

CNO & 
Given a hate speech and a counter narrative in response to the hate speech, determine how offensive the counter narrative is. \\

\hhline{--~}

CNSt & 
Evaluate the stance of the counter narrative given a hate speech and a counter narrative in response to the hate speech. \\

\midrule

HSO & 
You are given a hate speech. Determine how offensive the hate speech is. &
hate speech \texttt{<sep>} \textcolor{blue}{According to a recent OFSTED report, a school in Birmingham is still segregating girls and boys, despite a Hight Court ruling in 2017 that this is unlawful.}\\

\midrule

SRC & 
You have a story that consists of a premise, a second sentence, and an ending. A second story with the same premise is provided but the second sentence is replaced with a counterfactual sentence, and the ending is rewritten, determine how reasonable the rewritten ending is given the premise and counterfactual sentence provided. &
\multirow{20}{0.45\textwidth}{story \texttt{<sep>} Wallace had a passion for gardening. Wallace never became a professional gardener. \texttt{<sep>} ending \texttt{<sep>} \textcolor{blue}{One day, he was approached by a professional gardener. The professional gardener suggested that they work together. Now, Wallace has a lot of money that he loves to pay back.}}\\

\hhline{--~}

SRE & 
You have a story that consists of a premise, a second sentence, and an ending. A second story with the same premise is provided but the second sentence is replaced with a counterfactual sentence, and the ending is rewritten. Determine how reasonably the rewritten ending outlines a sequence of events.\\

\hhline{--~}

SRPl & 
You are given a story, an original ending and a rewritten ending. Determine how well the plot in the rewritten ending relates to the plot of the original ending.\\

\hhline{--~}

SRS & 
Determine how well the rewritten ending keeps in mind the details provided in the counterfactual given a story that consists of a premise, a second sentence, and an ending as well as a second story with the same premise is provided but the second sentence is replaced with a counterfactual sentence, and the ending is rewritten.\\

\hhline{--~}

SRPr & 
Determine how well the rewritten ending keeps in mind the details provided in the premise given a story that consists of a premise, a second sentence, and an ending as well as a second story with the same premise is provided but the second sentence is replaced with a counterfactual sentence, and the ending is rewritten.\\




\midrule

DIA & 
You are given a premise, a hypothesis, and an update sentence. Determine how much the much the update sentence weakens the hypothesis. &
A girl in a black sweater and jeans pours water into an empty soda bottle. \texttt{<sep>} A girl pours water into an empty coca cola bottle \texttt{<sep>} \textcolor{blue}{The bottle is empty}\\

\midrule

DII & 
Given a premise, a hypothesis, and an update sentence, determine how much the much the update sentence strengthens the hypothesis. &
A group of mountain climbers rests at the summit. \texttt{<sep>} A group of climbers celebrates at the top of Everest. \texttt{<sep>} \textcolor{blue}{The climbers are smiling}\\

\bottomrule
\end{tabular}
\caption{Natural language instructions used for each task alongside data samples.}
\label{tab:instructions}
\end{table*}

Table~\ref{tab:instructions} presents the natural language instructions used for each task, along with an example for each task.

\end{document}